\pdfoutput=1

\documentclass[11pt]{article}

\usepackage{amsmath,amsfonts,bm}

\def\eqref#1{equation~\ref{#1}}

\def\1{\bm{1}}

\def\vh{{\bm{h}}}

\def\vp{{\bm{p}}}
\def\vq{{\bm{q}}}

\def\mS{{\bm{S}}}

\DeclareMathAlphabet{\mathsfit}{\encodingdefault}{\sfdefault}{m}{sl}
\SetMathAlphabet{\mathsfit}{bold}{\encodingdefault}{\sfdefault}{bx}{n}

\usepackage[]{ACL2023}

\usepackage{times}
\usepackage{latexsym}
\usepackage{graphicx}

\usepackage{microtype}
\usepackage{multirow}
\usepackage{subfigure}
\usepackage{booktabs} 

\usepackage{xcolor}

\usepackage{dsfont}

\usepackage[T1]{fontenc}

\usepackage[utf8]{inputenc}

\usepackage{microtype}

\usepackage{inconsolata}

\title{GLiNER: Generalist Model for Named Entity Recognition using Bidirectional Transformer}

\author{Urchade Zaratiana$^{1,2}$, Nadi Tomeh$^2$, Pierre Holat$^{1,2}$, Thierry Charnois$^2$ \\
$^1$ FI Group, 
$^2$ LIPN, CNRS UMR 7030, France \\{\tt zaratiana@lipn.fr} \\ \texttt{https://github.com/urchade/GLiNER}}

\begin{document}

\maketitle

\begin{abstract}
Named Entity Recognition (NER) is essential in various Natural Language Processing (NLP) applications. Traditional NER models are effective but limited to a set of predefined entity types. In contrast, Large Language Models (LLMs) can extract arbitrary entities through natural language instructions, offering greater flexibility. However, their size and cost, particularly for those accessed via APIs like ChatGPT, make them impractical in resource-limited scenarios. In this paper, we introduce a compact NER model trained to identify any type of entity. Leveraging a bidirectional transformer encoder, our model, GLiNER, facilitates parallel entity extraction, an advantage over the slow sequential token generation of LLMs. Through comprehensive testing, GLiNER demonstrate strong performance, outperforming both ChatGPT and fine-tuned LLMs in zero-shot evaluations on various NER benchmarks. 
\end{abstract}

\section{Introduction}

Named Entity Recognition plays a crucial role in various real-world applications, such as constructing knowledge graphs. Traditional NER models are limited to a predefined set of entity types. Expanding the number of entity types can be beneficial for many applications but may involve labeling additional datasets. The emergence of Large Language Models, like GPT-3 \citep{Brown2020LanguageMA}, has introduced a new era for open-type NER by enabling the identification of any types of entity types only by natural language instruction. This shift signifies a significant departure from the inflexibility observed in traditional models. However, powerful LLMs typically consist of billions of parameters and thus require substantial computing resources. Although it is possible to access some LLMs via APIs \citep{openai2023gpt4}, using them at scale can incur high costs.

\begin{figure}[t]
    \centering
\includegraphics[width=1\columnwidth]{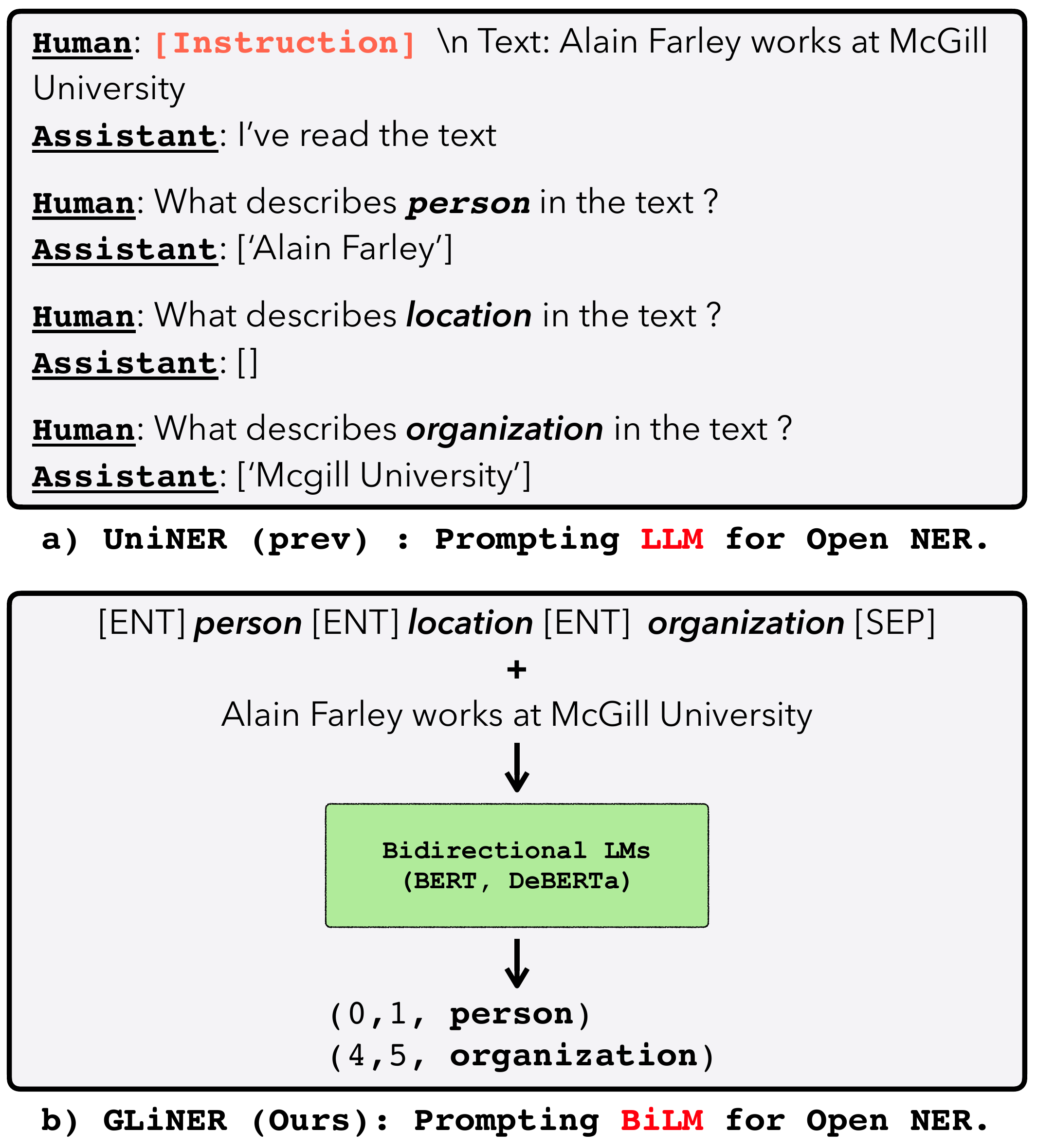}
    \caption{\textbf{BiLM for Open NER.} Previous models, such as UniNER \citep{zhou2023universalner} (Fig. a), approach the task of open type NER by prompting LLMs with natural language instructions (using a multi-turn dialogue style). Our proposed GLiNER utilizes small bidirectional LMs and treats the NER task as matching entity types with textual spans in a latent space.}
    \label{fig:throu}
\end{figure}

\begin{figure*}[t]
    \centering
\includegraphics[width=0.999\textwidth]{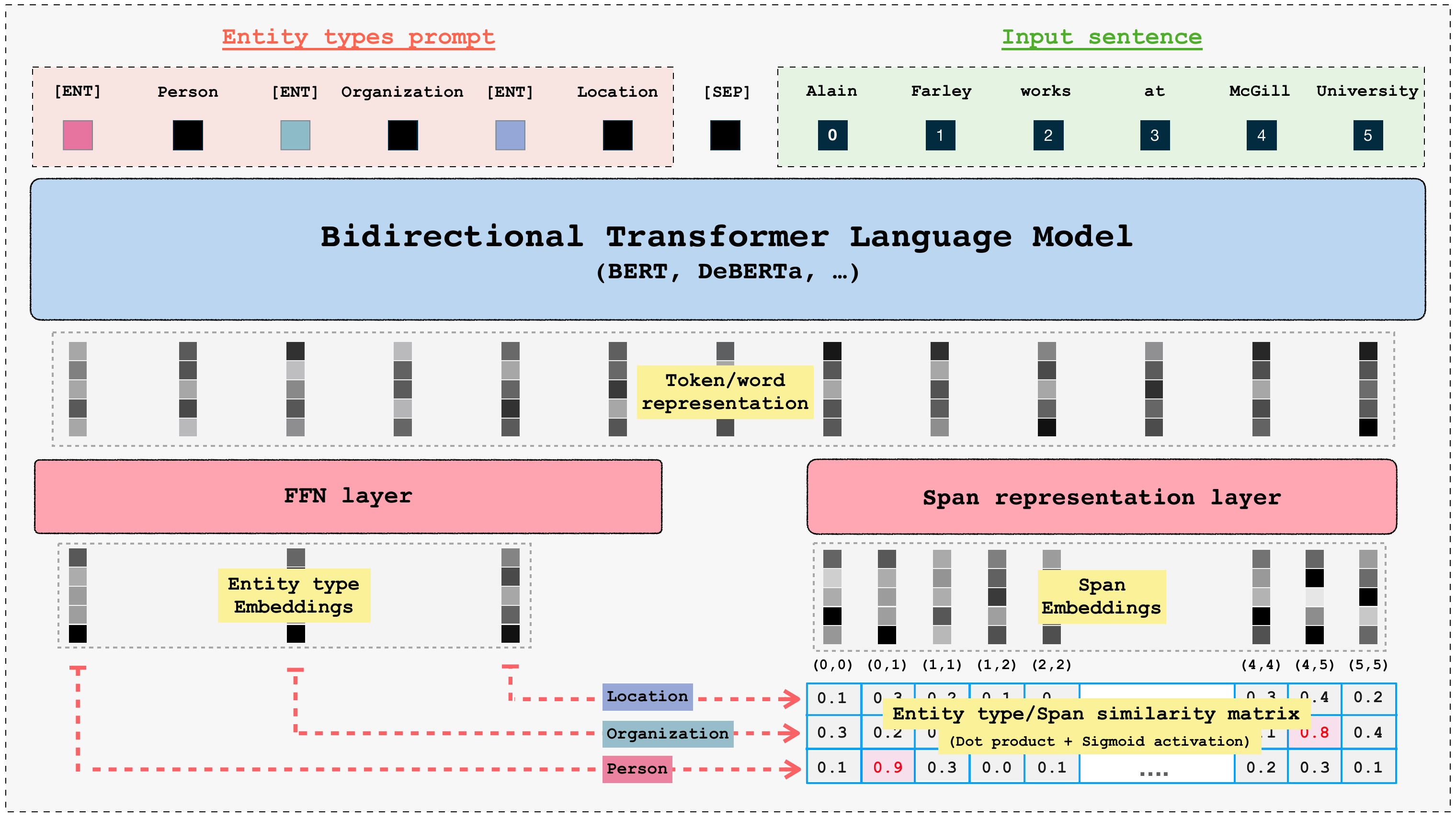}
    \caption{\textbf{Model architecture}. GLiNER employs a BiLM and takes as input entity type prompts and a sentence/text. Each entity is separated by a learned token \texttt{[ENT]}. The BiLM outputs representations for each token. Entity embeddings are passed into a FeedForward Network, while input word representations are passed into a span representation layer to compute embeddings for each span. Finally, we compute a matching score between entity representations and span representations (using dot product and sigmoid activation). For instance, in the figure, the span representation of (0, 1), corresponding to "Alain Farley," has a high matching score with the entity embeddings of "Person".}
    \label{fig:main}
\end{figure*}

Recently, researchers have explored the fine-tuning of open source language models such as LLaMa \citep{Touvron2023LLaMAOA} for named entity recognition tasks. \citet{Wang2023InstructUIEMI}, for example, introduced InstructUIE, a fine-tuned FlanT5-11B \citep{Raffel2019ExploringTL,chung2022scaling} model on existing information extraction datasets, achieving excellent performance in zero-shot settings. Additionally, GoLLIE \citep{Sainz2023GoLLIEAG} was introduced as an extension of InstructUIE work by by finetuning a CodeLLaMa \citep{Rozire2023CodeLO} using detailed annotation guidelines, resulting in significant performance improvements. Another recent proposal by \citet{zhou2023universalner}, called UniversalNER, involves the fine-tuning of LLMs using diverse data from various domains annotated with ChatGPT instead of relying on standard NER datasets. Their approach enables the replication and even surpassing of the original capability of ChatGPT when evaluated in zero-shot on several NER datasets. While these works have achieved remarkable results, they present certain limitations we seek to address: They use autoregressive language models, which can be slow due to token-by-token generation; Moreover, they employ large models with several billion parameters, limiting their deployment in compute-limited scenarios. Furthurmore, as NER is treated as a text generation problem, the generation of entities is done in several decoding steps, and there is no way to perform the prediction of multiple entity types in parallel. 

In our work, we propose a model that addresses the above-mentioned problems. Instead of relying on large autoregressive models, we utilize smaller-scale Bidirectional Language Models (BiLM), such as BERT \citep{Devlin2019BERTPO} or deBERTa \citep{He2021DeBERTaV3ID}. The core concept of our model involves treating the task of Open NER as matching entity type embeddings to textual span representations in latent space, rather than as a generation task.  This approach naturally solves the scalability issues of autoregressive models and allows for bidirectional context processing, which enables richer representations. When trained on the dataset released by \citep{zhou2023universalner}, which comprises texts from numerous domains and thousands of entity types, our model demonstrates impressive zero-shot performance. More specifilcally, it outperforms both ChatGPT and fine-tuned LLMs on zero-shot NER datasets (Table \ref{tab:ood}). Our model's robustness is further evident in its ability to handle languages not even encountered during training. Specifically, our model surpasses ChatGPT in 8 of 10 languages (Table \ref{crossling})  that were not included in its training data.

\section{Method}
This section presents our model, GLiNER, which is train to extract any types of entity  using a Bidirectional Language Models. Our model has three main components: i) a pre-trained textual encoder (a BiLM such as BERT), ii) a span representation module which computes span embeddings from token embeddings, iii) an entity representation module which computes entity embeddings that the model seeks to extract. The goal is to have entity and span embeddings in the same latent space to assess their compatibility (degree of matching). The overall architecture of our model is depicted in Figure \ref{fig:main}.

\subsection{Architecture}

\paragraph{Input format} The input to our model comprises a unified sequence combining entity types (expressed in natural language) and the input text from which entities are to be extracted. The input format is as follows:

\begin{figure}[h]
    \centering
    \includegraphics[width=1\columnwidth]{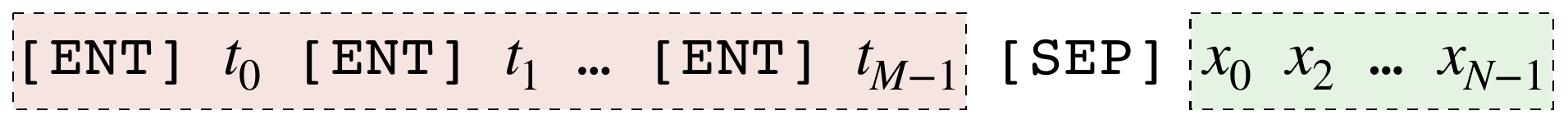}
\end{figure}

\texttt{[ENT]} token represents a special token placed before each entity type and the \texttt{[SEP]} token functions as a delimiter, separating the sequence of entity types from the input text. They are initialized randomly at the start of training. 

\paragraph{Token representation} The token encoder processes the unified input to compute interactions between all tokens (both entity types and input text), producing contextualized representations. Let $\vp=\{\vp_i\}_0^{M-1} \in \mathbb{R}^{M\times D}$ represent the encoder's output for each entity type, corresponding to all the \texttt{[ENT]} token representations. Similarly, $\vh=\{\vh_i\}_0^{N-1} \in \mathbb{R}^{N\times D}$ denotes the representation of each word in the input text. For words tokenized into multiple subwords, we use the representation of the first subword, which is a standard choice in the NER literature \citep{zaratiana-etal-2022-named}.

\paragraph{Entity and Span Representation} In our model, we aim to encode entity types and span embeddings into a unified latent space. The entity representation is computed by refining the initial representation $\vp$ using a two-layer feedforward network, resulting in $\vq=\{\vq_i\}_0^{M-1} \in \mathbb{R}^{M\times D}$. The representation of a span starting at position $i$ and ending at position $j$ in the input text, $\mS_{ij} \in \mathbb{R}^{D}$, is computed as:
\begin{equation}
\mS_{ij} = \texttt{FFN}(\vh_i \otimes \vh_j)
\end{equation}

Here, \texttt{FFN} denotes a two-layer feedforward network, and $\otimes$ represents the concatenation operation. In practice, The computation of all span representations can be easily parallelized. Moreover, we set an upper bound to the length (K=12) of the span in order to keep linear complexity, without harming recall.

\paragraph{Entity Type and Span Matching}
To evaluate whether a span $(i, j)$ corresponds to entity type $t$, we calculate the following matching score:

\begin{equation}
\phi(i,j,t) = \sigma(\mS_{ij}^T \vq_t) \in \mathbb{R}
\end{equation}

In this equation, $\sigma$ denotes a sigmoid activation function. As we train with binary cross-entropy loss (see next sec. \ref{train}), $\phi(i,j,t)$ can be interpreted as the probability of the span $(i,j)$ being of type $t$.

\begin{table*}[h]
\centering
{\fontsize{10pt}{12pt}\selectfont
\begin{tabular}{l|l|cccccccc}
\toprule
Model  & Params & Movie & Restaurant & AI & Literature & Music & Politics & Science & Average \\
\midrule
Vicuna-7B & 7B & 6.0 & 5.3 & 12.8 & 16.1 & 17.0 & 20.5 & 13.0 & 13.0 \\
Vicuna-13B & 13B & 0.9 & 0.4 & 22.7 & 22.7 & 26.6 & 27.0 & 22.0 & 17.5 \\
USM & 0.3B & 37.7 & 17.7 & 28.2 & 56.0 & 44.9 & 36.1 & 44.0 & 37.8 \\
ChatGPT & -- & 5.3 & 32.8 & 52.4 & 39.8 & 66.6 & 68.5 & \textbf{67.0} & 47.5 \\
InstructUIE & 11B & \textbf{63.0} & 21.0 & 49.0 & 47.2 & 53.2 & 48.1 & 49.2 & 47.2  \\
UniNER-7B & 7B & 42.4 & 31.7 & 53.6 & 59.3 & 67.0 & 60.9 & 61.1 & 53.7 \\
UniNER-13B & 13B & 48.7 & 36.2 & 54.2 & 60.9 & 64.5 & 61.4 & 63.5 & 55.6 \\
GoLLIE & 7B & \textbf{63.0} & \textbf{43.4} & \textbf{59.1} & 62.7 & 67.8 & 57.2 & 55.5 & 58.0 \\
\midrule
GLiNER-S & 50M & 46.9 & 33.3 & 50.7 & 60.0 & 60.9 & 61.5 & 55.6 & 52.7 \\
GLiNER-M & 90M & 42.9 & 37.3 & 51.8 & 59.7 & 69.4 & 68.6 & 58.1 & 55.4 \\
GLiNER-L & 0.3B & 57.2 & 42.9 & 57.2 & \textbf{64.4} & \textbf{69.6} & \textbf{72.6} & 62.6 & \textbf{60.9} \\
\bottomrule
\end{tabular}
}
\caption{\textbf{Zero-Shot Scores on Out-of-Domain NER Benchmark.} We report the performance of GLiNER with various DeBERTa-v3 \citep{He2021DeBERTaV3ID} model sizes. Results for Vicuna, ChatGPT, and UniNER are from \citet{zhou2023universalner}; USM and InstructUIE are from \citet{Wang2023InstructUIEMI}; and GoLLIE is from \citet{Sainz2023GoLLIEAG}.}
\label{tab:ood}
\end{table*}

\subsection{Training \label{train}}

During training, our objective is to optimize model parameters to enhance the matching score for correct span-type pairs (positive pairs) and reduce it for incorrect pairs (negative pairs). A span $(i, j)$ paired with an entity type $t$ forms a positive pair ($s \in \mathcal{P}$) if the span is labeled with type $t$ in the training data. Otherwise, it is a negative pair ($s \in \mathcal{N}$). The training loss for an individual example, comprising spans $\mathcal{S}$ and entity types $\mathcal{T}$, is defined as:

\begin{equation}
\begin{aligned}
\mathcal{L}_{\text{BCE}} = -\sum_{s \in \mathcal{S} \times \mathcal{T}} &\mathbb{I}_{s \in \mathcal{P}} \log \phi(s) + \\ &\mathbb{I}_{s \in \mathcal{N}} \log \left(1 - \phi(s)\right)
\end{aligned}
\end{equation}

The variable $s$ represents a pair of span/entity type and $\mathbb{I}$ is an indicator function, which returns 1 when the specified condition is true and 0 otherwise. This loss function corresponds to binary cross-entropy.

\subsection{Decoding algorithm}

In the decoding phase, we employ a greedy span section \citep{zaratiana-etal-2022-named} that selects entity spans based on matching scores, to ensure task/dataset specific constraints. This strategy is applied independently to each sentence. Only, spans \( (i, j) \) with matching scores \( \phi(i, j, c) > 0.5 \) are considered for selection.

\paragraph{Flat NER:}  The algorithm chooses the highest-scoring non-overlapping span and continues this process until all spans are evaluated.

\paragraph{Nested NER:} Similar to Flat NER, but the algorithm allows selection of fully nested spans within other entities while still avoiding partial overlaps.

\paragraph{Algorithm Efficiency:} The decoding is implemented using a priority queue for spans, ensuring an \( O(n \log n) \) complexity, with \( n \) being the number of candidate spans.

\section{Experimental Setting}
\subsection{Training data}
Our objective is to construct a versatile NER model capable of accurately identifying a wide array of entity types across different textual domains. To achieve this, it is essential that our training dataset encompasses a diverse range of entity types. For this, we utilize the training data released by \citet{zhou2023universalner}, known as Pile-NER\footnote{https://huggingface.co/datasets/Universal-NER/Pile-NER-type}. This dataset is derived from the Pile corpus \citep{Gao2020ThePA}, commonly used for pretraining large language models, and comprises text from diverse sources. More specifically, \citet{zhou2023universalner} sampled 50,000 texts from the Pile data and employed ChatGPT to extract their associated entity types. Notably, they did not specify the entity types to the LLMs, aiming to extract a diverse range of entity types. They used the following prompt:
\begin{figure}[h]
    \centering
    \includegraphics[width=1\columnwidth]{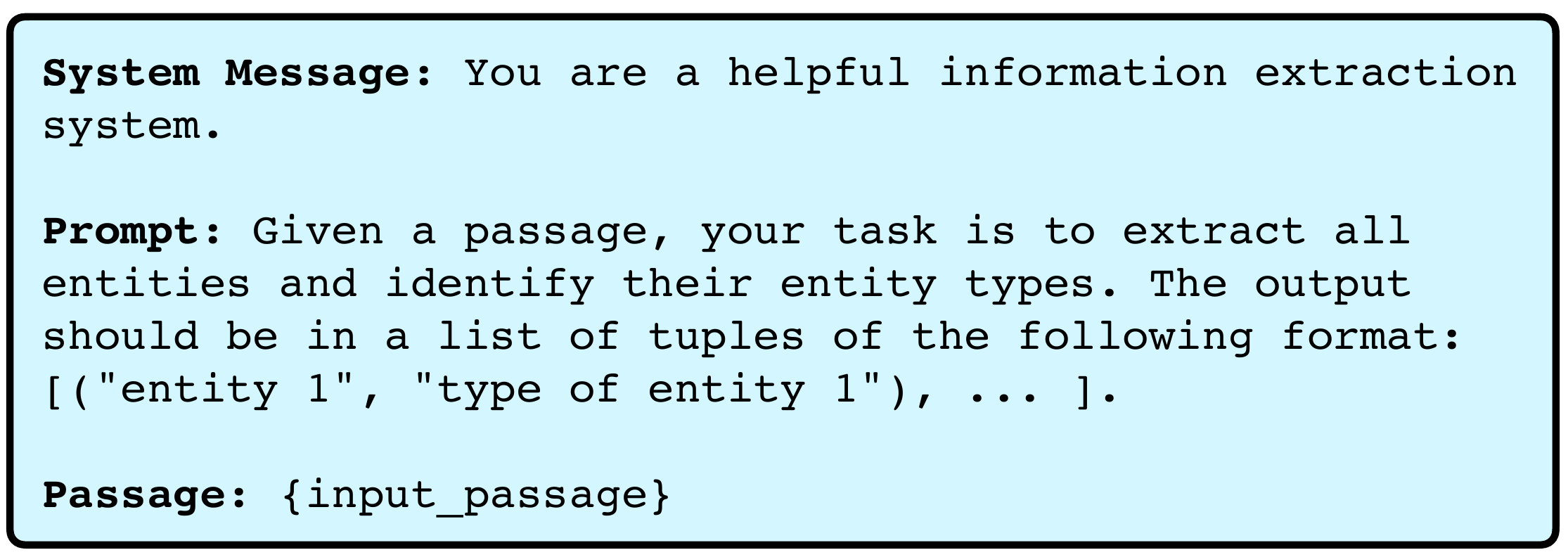}
    \caption{\textbf{Prompting ChatGPT for entity extraction}. This prompt was used \citet{zhou2023universalner} to construct the Pile-NER dataset.}
    \label{fig:promp}
\end{figure}

Finally, after filtering bad outputs their datasets results in 44889 passages containing in total 240k entity spans and 13k distinct entity types.

\subsection{Hyperparameters \label{hyper}} 

Our model, GLiNER, is trained on the Pile-NER dataset, which we described in the previous section. We use the deBERTa-v3 \citep{He2021DeBERTaV3ID} as our backbone due to its proven empirical performance. All non-pretrained layers have a width dimension of 768 and a dropout rate of 0.4. Regarding the training process, we employ the AdamW optimizer \citep{Loshchilov2017DecoupledWD}, setting a base learning rate of 1e-5 for pretrained layers (the transformer backbone) and 5e-5 for non-pretrained layers (FFN layers, span representation). We trained our models for a maximum of 30k steps, starting with a 10\% warmup phase, followed by a decay phase using a cosine scheduler.
The Pile-NER dataset natively contains only positive entities (i.e., entities that are present in the sentence), and we found it useful to include negative entity types during training. This is achieved by sampling random entities from other examples in the same batch. In addition, we follow the strategies outlined in \citet{Sainz2023GoLLIEAG} as a form of regularization, which includes \textit{shuffling entity order} and \textit{randomly dropping entities}. Furthermore, we limit the number of entity types to 25 per sentence during training. The larger variant of our model, GLiNER-L, takes 5 hours to train on an A100 GPU.

\subsection{Baselines}

In our evaluation, we compare our model, GLiNER, with several recent models designed for open-type NER. First, we examine chat models like \textbf{ChatGPT} and \textbf{Vicuna} \citep{vicuna2023}, which utilize the prompting from \citet{Ye2023ACC}; we report their results reported by \citet{zhou2023universalner}. Next, we focus on three recent Large Language Models (LLMs) that have been fine-tuned for NER: \textbf{InstructUIE} \citep{Wang2023InstructUIEMI}, based on the FlanT5 11B model and fine-tuned on various NER datasets; \textbf{UniNER} \citep{zhou2023universalner}, which employs a LLaMa model fine-tuned on a dataset generated by ChatGPT; \textbf{GoLLIE} \citep{Sainz2023GoLLIEAG}, fine-tuned to adhere to detailed annotation guidelines for enhanced performance in unseen IE tasks, utilizing CodeLLama as its base model. Finally, we include \textbf{USM} \citep{Lou2023UniversalIE} in our comparison, which is similar in size to ours but features a different architecture.

\subsection{Evaluation}

\paragraph{Datasets} We primarily evaluate our model in a zero-shot context on common NER benchmarks, following previous works \cite{Wang2023InstructUIEMI,zhou2023universalner}. The first is the \textit{OOD NER Benchmark} (Table \ref{tab:ood}), which comprises seven diverse NER datasets from CrossNER and MIT. This benchmark is typically used for evaluating out-of-domain generalization capabilities of NER models. The second benchmark consists of \textit{20 NER datasets} (Table \ref{20ner}) from a wide range of domains, including biomedical, news articles, and tweets. These datasets are commonly used for training supervised NER models. Additionally, we evaluate our model on multilingual NER datasets (Table \ref{crossling}) for further investigation. For this purpose, we use the recently released \textit{Multiconer} (Multilingual Complex NER) \citep{Malmasi2022MultiCoNERAL}, which contains data in 11 languages across various domains.

\paragraph{Metric}  We adopt the standard NER evaluation methodology, calculating F1-score based on the exact match between predicted and actual entities.

\begin{table}[]
\centering
{\fontsize{8pt}{10pt}\selectfont
\begin{tabular}{lccc}
\toprule
\textbf{Dataset} & \textbf{ChatGPT} & \textbf{UniNER-7B} & \textbf{GLiNER-L} \\
\midrule
ACE05 & 26.6 & \textbf{36.9} & 27.3 \\
AnatEM & 30.7 & 25.1 & \textbf{33.3} \\
bc2gm & 40.2 & 46.2 & \textbf{47.9} \\
bc4chemd & 35.5 & \textbf{47.9} & 43.1 \\
bc5cdr & 52.4 & \textbf{68.0} & 66.4 \\
Broad Tweeter & 61.8 & \textbf{67.9} & 61.2 \\
CoNLL03 & 52.5 & \textbf{72.2} & 64.6 \\
FabNER & 15.3 & \textbf{24.8} & 23.6 \\
FindVehicle & 10.5 & 22.2 & \textbf{41.9} \\
GENIA & 41.6 & 54.1 & \textbf{55.5} \\
HarveyNER & 11.6 & 18.2 & \textbf{22.7} \\
MIT Movie & 5.3 & 42.4 & \textbf{57.2} \\
MIT Restaurant & 32.8 & 31.7 & \textbf{42.9} \\
MultiNERD & 58.1 & 59.3 & \textbf{59.7} \\
ncbi & 42.1 & 60.4 & \textbf{61.9} \\
OntoNotes & 29.7 & 27.8 & \textbf{32.2} \\
PolyglotNER & 33.6 & 41.8 & \textbf{42.9} \\
TweetNER7 & 40.1 & \textbf{42.7} & 41.4 \\
WikiANN & 52.0 & 55.4 & \textbf{58.9} \\
WikiNeural & 57.7 & 69.2 & \textbf{71.8} \\
\midrule
\textbf{Average} & 36.5 & 45.7 & \textbf{47.8} \\
\bottomrule
\end{tabular}
}
\caption{Z\textbf{ero-shot performance on 20 NER datasets.} Results of ChatGPT and UniNER are reported from \citep{zhou2023universalner}.}
\label{20ner}
\end{table}

\section{Results}

\subsection{Zero-shot on English datasets} 
In this section, we discuss the performance of our model in a zero-shot context, i.e., by only training on the Pile-NER dataset without further fine-tuning on target datasets.

\paragraph{OOD NER Benchmark} We first evaluate our model on the OOD benchmark as reported in Table \ref{tab:ood}. We compare three different sizes of our model (small, medium, and large) against the baselines. The results demonstrate our model's impressive capability, irrespective of its size. For example, even our smallest model, with only 50M parameters, outperforms general-purpose models such as ChatGPT and Vicuna. It also shows better performance than the 11B InstructUIE, which has been instruction-tuned for the NER task. Furthermore, when compared to UniNER, which used the same training data as GLiNER, our medium-sized model (90M) achieves comparable results to UniNER-13B (55 F1 for both), despite being 140 times smaller, while our largest version outperforms it by an average of 5 points. Our most best competitor, GoLLIE, which leads among the LLMs, achieves better performance than most of our models but is still less effective than GLiNER-L. Against USM, which has a comparable number of parameters to ours, the performance is significantly lower, highlighting the superiority of our architecture.

\paragraph{20 NER Benchmark} Table 2 presents a comparison of our model against ChatGPT and UniNER across 20 diverse NER datasets. First, similar to the OOD benchmark, ChatGPT significantly lags behind fine-tuned models for NER, trailing behind UniNER. Furthermore, GLiNER achieves the highest performance on 13 of these datasets, surpassing UniNER by an average of 2 points. This superior performance underscores GLiNER's robustness and adaptability across a broad spectrum of domains. However, a notable observation is that GLiNER underperforms compared to UniNER on tweet-based NER datasets. This highlights potential areas for improvement in GLiNER’s ability to process informal, colloquial, or noisy data, typical of social media content.

\begin{table}[]
  \centering
  {\fontsize{8pt}{10pt}\selectfont
\begin{tabular}{cl|c|ccc}
    \toprule
    & \multirow{2}{*}{\textbf{Language}} & \multirow{2}{*}{\textbf{Sup.}} & \multirow{2}{*}{\textbf{ChatGPT}} & \multicolumn{2}{c}{\textbf{GLiNER}} \\
    &          &      &         & En    & Multi \\
    \midrule
    \multirow{4}{*}{\rotatebox{90}{Latin}} 
    & German   & 64.6 & 37.1    & 35.6  & \textbf{39.5} \\
    & English  & 62.7 & 37.2    & \textbf{42.4}  & 41.7 \\
    & Spanish  & 58.7 & 34.7    & 38.7  & \textbf{42.1} \\
    & Dutch    & 62.6 & 35.7    & 35.6  & \textbf{38.9} \\
    \midrule
    \multirow{7}{*}{\rotatebox{90}{Non-Latin}}
    & Bengali  & 39.7 & 23.3    & 0.89  & \textbf{25.9} \\
    & Persian  & 52.3 & 25.9    & 14.9  & \textbf{30.2} \\
    & Hindi    & 47.8 & 27.3    & 11.3  & \textbf{27.8} \\
    & Korean   & 55.8 & \textbf{30.0}    & 20.5  & 28.7 \\
    & Russian  & 59.7 & 27.4    & 30.3  & \textbf{33.3} \\
    & Turkish  & 46.8 & \textbf{31.9}    & 22.0  & 30.0  \\
    & Chinese  & 53.1 & 18.8    & 6.59  & \textbf{24.3}   \\
    \midrule
    \multicolumn{2}{l|}{\textbf{Average}} & 54.9 & 29.9 & 23.6 & \textbf{32.9} \\
    \bottomrule
  \end{tabular}
  }
  \caption{\textbf{Zero-Shot Scores on Different Languages.} The baseline, \textbf{Sup.}, is an XLM-R \citep{Conneau2019UnsupervisedCR} model fine-tuned on the training set of each language separately, as reported by \citet{Malmasi2022MultiCoNERAL}. ChatGPT evaluation is taken from \citet{Lai2023ChatGPTBE}. GLiNER-En employs deBERTa-v3-Large, and Multi uses mdeBERTa-v3-base.}
  \label{crossling}
\end{table}

\subsection{Zero-Shot Multilingual Evaluation}
In this section, we evaluate the performance of our model in a zero-shot context on unseen languages to assess its generalizability. This evaluation uses the Multiconel dataset, with results detailed in Table 3. Our model, GLiNER, is presented in two variants: \textbf{En}, which employs deBERTa-v3-Large as its backbone, and \textbf{Multi}, which utilizes a multilingual version of deBERTa-v3 (mdeBERTa-v3). Both versions were fine-tuned on the Pile-NER dataset. For comparative purposes, we include results from ChatGPT and a supervised baseline, the latter being fine-tuned on the training set of each dataset using separate models.

\paragraph{Results}
As expected, the supervised baseline demonstrated superior performance, significantly outperforming the zero-shot models. Among these, GLiNER-Multi showed the most promising results, surpassing ChatGPT in most languages, which is noteworthy considering that the fine-tuning dataset, Pile-NER, consists exclusively of English examples. Interestingly, its performance on Spanish data slightly exceeded that in English. While GLiNER-En generally underperformed compared to ChatGPT on average, it achieved competitive, and at times superior, results in languages using the Latin script, such as Spanish and German. However, its performance was markedly less competitive in non-Latin languages, particularly in Bengali, where it scored only 0.89 in the F1 score.

\begin{table}[]
\centering
{\fontsize{8pt}{10pt}\selectfont
\begin{tabular}{lccccc}
\hline
\toprule
\multirow{2}{*}{\textbf{Dataset}} & \textbf{InstructUIE} & \textbf{UniNER-7B} & \multicolumn{2}{c}{\textbf{GLiNER-L}} \\ 
 & \textbf{w/o} & \textbf{w/} & \textbf{w/} & \textbf{w/o} \\ 
\midrule

ACE05 & 79.9 & \textbf{86.7} & 82.8 & 81.3 \\
AnatEM & 88.5 & 88.5 & \textbf{88.9} & 88.4 \\
bc2gm & 80.7 & 82.4 & \textbf{83.7} & 82.0 \\
bc4chemd & 87.6 & \textbf{89.2} & 87.9 & 86.7 \\
bc5cdr & 89.0 & \textbf{89.3} & 88.7 & 88.7 \\
Broad Twitter & 80.3 & 81.2 & 82.5 & \textbf{82.7} \\
CoNLL03 & 91.5 & \textbf{93.3} & 92.6 & 92.5 \\
FabNER & 78.4 & \textbf{81.9} & 77.8 & 74.8 \\
FindVehicle & 87.6 & \textbf{98.3} & 95.7 & 95.2 \\
GENIA & 75.7 & 77.5 & \textbf{78.9} & 77.4 \\
HarveyNER & \textbf{74.7} & 74.2 & 68.6 & 67.4 \\
MIT Movie & 89.6 & \textbf{90.2} & 87.9 & 87.5 \\
MIT Restaurant & 82.6 & 82.3 & \textbf{83.6} & 83.3 \\
MultiNERD & 90.3 & 93.7 & \textbf{93.8} & 93.3 \\
ncbi & 86.2 & 87.0 & \textbf{87.8} & 87.1 \\
OntoNotes & 88.6 & \textbf{89.9} & 89.0 & 88.1 \\
PolyglotNER & 53.3 & \textbf{65.7} & 61.5 & 60.6 \\
TweetNER7 & \textbf{65.9} & 65.8 & 51.4 & 50.3 \\
WikiANN & 64.5 & \textbf{84.9} & 83.7 & 82.8 \\
wikiNeural & 88.3 & \textbf{93.3} & 91.3 & 91.4 \\
\midrule
\textbf{Average} & 81.2 & \textbf{84.8} & 82.9 & 82.1 \\
\bottomrule
\end{tabular}
}
\caption{\textbf{In-domain Supervised Finetuning.} All the models are fine-tuned on the mix of all training data of the benchmark. \textbf{w/} indicates that the model was trained on the Pile-NER dataset before finetuning.}
\label{tab:supexp}
\end{table}

\subsection{In-domain Supervised tuning}
In our work, we also perform in-domain supervised fine-tuning (on 20 NER datasets) of our model to compare its capabilities against LLMs under this setup. Specifically, we compare our model against InstructUIE and UniNER, both of which have also been fine-tuned. The main difference is that UniNER has been pre-trained on the Pile-NER dataset before fine-tuning.

\paragraph{Training Setup} For the supervised setting, we primarily adhere to the same experimental setup as described in the main experiment (using deBERTa-v3 large), except for the training dataset. Regarding the training data, we follow the approach of InstructUIE: we randomly sample 10,000 data points for each dataset in the 20 NER benchmark. If a dataset does not contain 10,000 samples, we include all available data. We implement two variants of our model: the first one initializes the weights from our zero-shot model, which is a pretrained on the Pile-NER dataset. The second variant is trained without the Pile-NER dataset, same as InstructUIE.

\paragraph{Results} The results of our experiment are reported in Table 3. Firstly, we observe that for the in-domain fine-tuning, our GLiNER model, pretrained on Pile-NER, achieves slightly better results than the non-pretrained variant, with an average difference of 0.8. Moreover, our pretrained GLiNER model outperforms InstructUIE (with an average difference of 0.9) despite being fine-tuned on the same dataset, whereas InstructUIE is significantly larger (approximately 30 times so). This demonstrates that our proposed architecture is indeed competitive. However, our model falls behind UniNER by almost 3 points. Nevertheless, our model still manages to achieve the best score in 7 out of 20 datasets.

\begin{figure}
    \centering
    \includegraphics[width=.8\columnwidth]{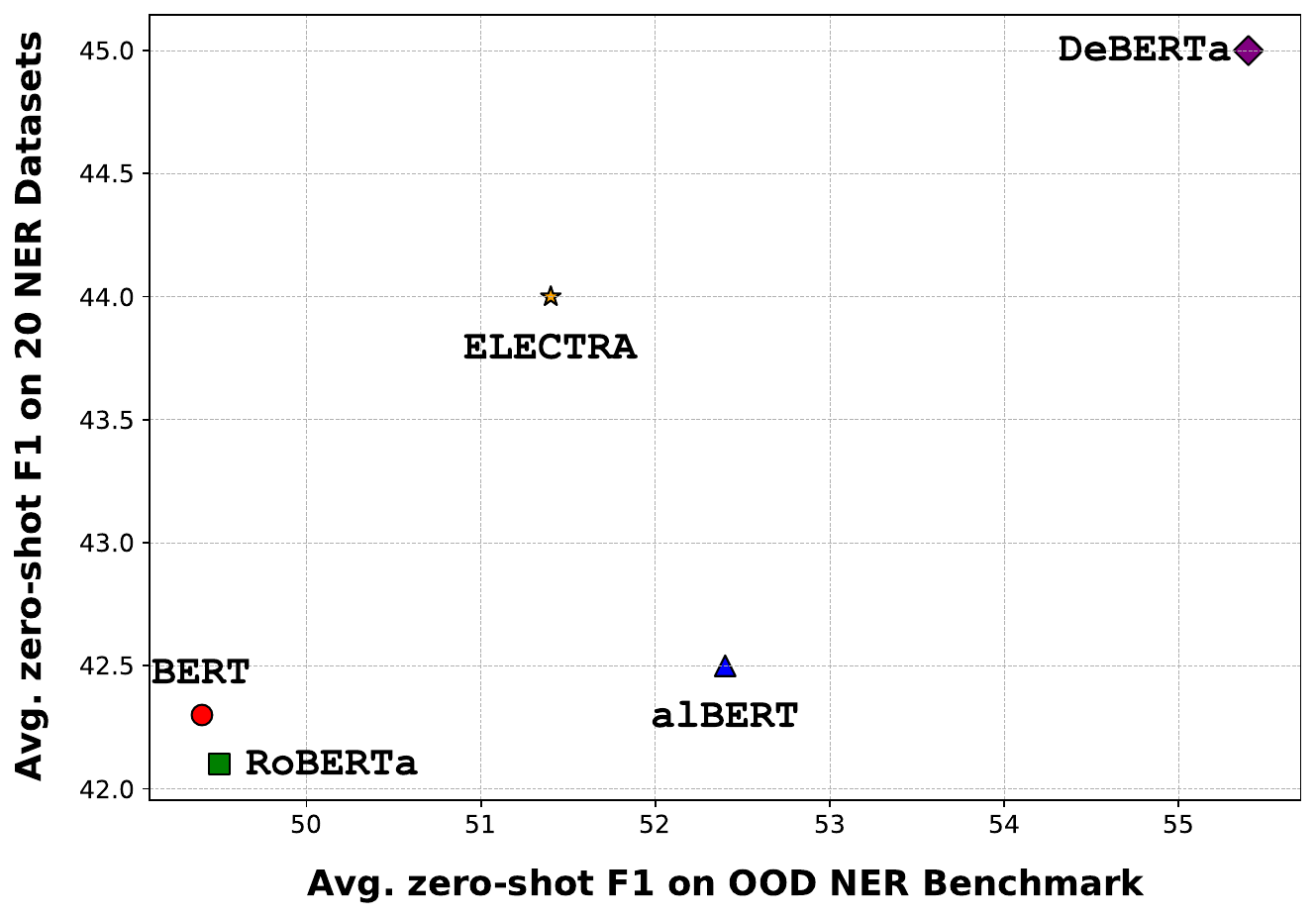}
    \caption{\textbf{Zero-shot performance for different backbones.} It reports the avg. results on 20 NER and OOD NER datasets}
    \label{fig:backbone}
\end{figure}

\section{Further analysis and ablations}
Here we conduct different set of experiments to better investigate our model.

\subsection{Effect of Different Backbones}
In our work, we primarily utilize the deBERTa-v3 model as our backbone due to its strong empirical performance. However, we demonstrate here that our method is adaptable to a wide range of BiLMs.

\paragraph{Setup}
Specifically, we investigate the performance of our model using other popular BiLMs, including BERT \citep{Devlin2019BERTPO}, RoBERTa \citep{Liu2019RoBERTaAR}, AlBERT \citep{Lan2019ALBERTAL}, and ELECTRA \citep{clark2020electra}. We also conducted experiments with XLNet \citep{Yang2019XLNetGA} but did not achieve acceptable performance (achieving at most 3 F1 on the OOD benchmark) despite extensive hyperparameter tuning. For a fair comparison, we employed the base size (GLiNER-M) and tuned the learning rate for each model. We report the zero-shot results on both the OOD benchmark and the 20 NER benchmark in Figure \ref{fig:backbone}.

\paragraph{Result} The results of our experiment, as shown in the Figure \ref{fig:backbone}, clearly demonstrate the superiority of deBERTa-v3 over other pretrained BiLMs. It achieves the highest performance on both benchmarks by a clear margin. ELECTRA and AlBERT also show notable performance, albeit slightly lower, while BERT and RoBERTa lag behind with similar scores. However, it should be noted that all of the backbones we tested demonstrate strong performance compared to existing models. More specifically, even BERT-base, which ranks among the lower performers, achieves around 49 F1 on the OOD benchmark. This score is still 2 F1 points higher than the average for models like ChatGPT and InstructUIE.

\subsection{Effect of Pretraining on In-domain Performance}

\begin{figure}
    \centering
    \includegraphics[width=0.8\columnwidth]{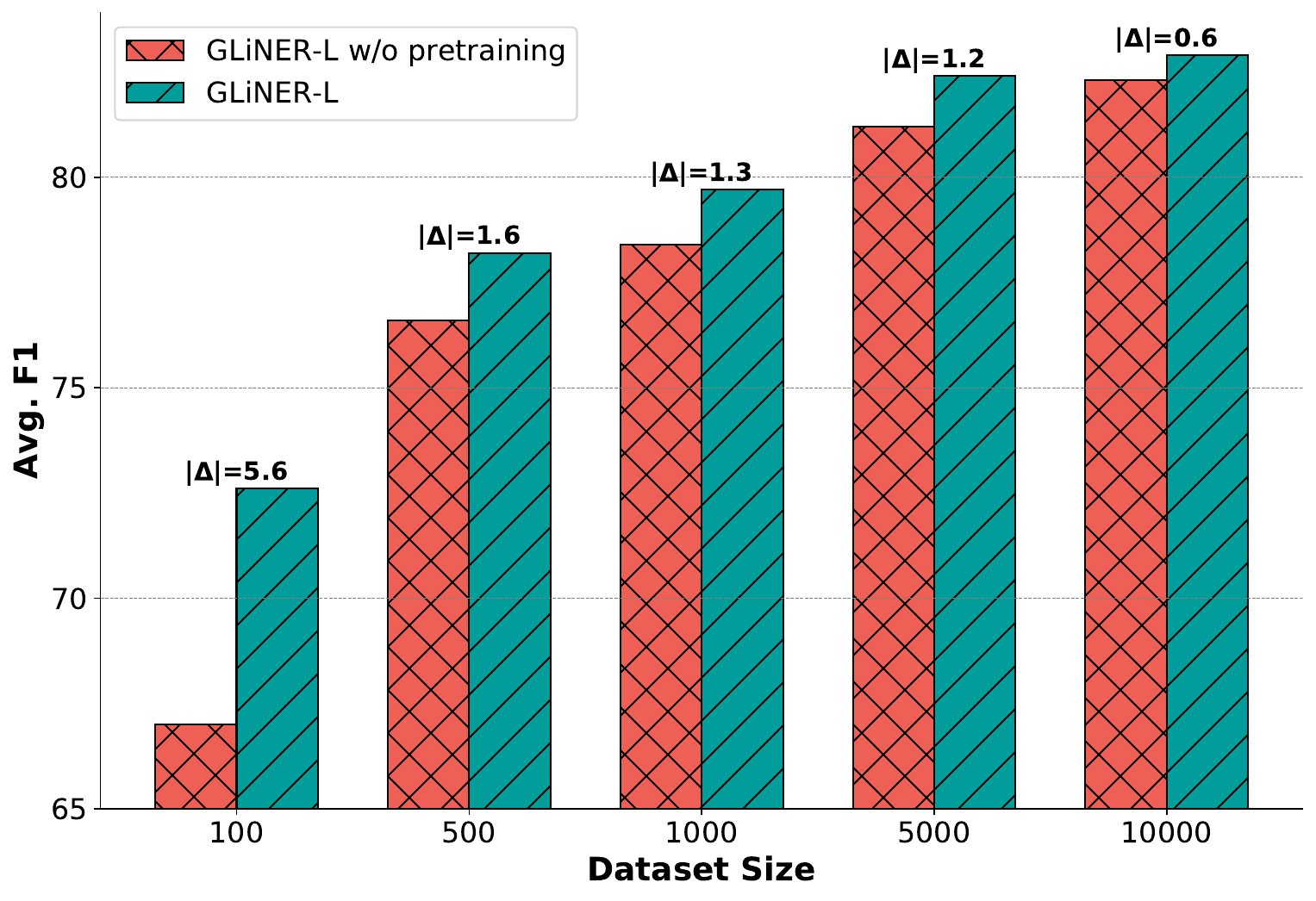}
    \caption{\textbf{Supervised performance across different dataset sizes.} The evaluation is conducted on the 20 NER datasets (in table \ref{tab:supexp}).}
    \label{fig:num-data}
\end{figure}

In this section, we investigate the impact of pretraining on the Pile-NER dataset for supervised in-domain training on the 20 NER datasets, across various data sizes. The experiments range from 100 samples per dataset to 10,000 (full training setup). We use the same hyperparameters for all configurations. The results are reported in Figure \ref{fig:num-data}.

\paragraph{Results} As shown in the figure, models pretrained on Pile-NER consistently outperform their counterparts that are only trained on supervised data, indicating successful positive transfer. We further observe that the gain is larger when supervised data is limited. For instance, the difference in performance is 5.6 when employing 100 samples per dataset, and the gap becomes smaller as the size of the dataset increases.

\begin{table}[h]
\centering
\label{tab:experiment_results}
  {\fontsize{10}{13pt}\selectfont
\begin{tabular}{@{}lccc@{}}
\toprule
\textbf{Negative Samples} & \textbf{Prec} & \textbf{Rec} & \textbf{F1} \\ \midrule
0\%  & \textcolor{red}{49.3} & 58.1 & 53.3 \\
50\% & \textbf{62.3} & \textbf{59.7} & \textbf{60.9} \\
75\% & 61.1 & \textcolor{red}{56.5} & 58.6 \\ \bottomrule
\end{tabular}
}
\caption{\textbf{Effect of negative entity types sampling.}}
\label{negsamp}
\end{table}

\subsection{Ablations}

\paragraph{Negative Entity Sampling}
 The original Pile-NER dataset, curated by \citet{zhou2023universalner}, features passages with positive entity instances, i.e., entities that are directly present in the text. To better align training with real-world scenarios, where some entity types might be absent, we implemented negative entity sampling as mentioned in Section \ref{hyper}. In this study, we evaluate different sampling ratios: 0\% (only positive entities), 50\%, and 75\% negative entities. 
 Table \ref{negsamp} shows that training with only positive entities results in lower precision but higher recall, indicating that the model often makes false positive errors. Conversely, using 75\% negative entities increases precision but decreases recall, as the abundance of negatives makes the model more cautious, leading to missed correct entities. A 50\% negative entity ratio proves to be the most effective, providing a balanced approach.

\paragraph{Entity type dropping}
In the experiment, we employed a strategy of randomly varying the number of entity prompts during training. This approach aimed to expose the model to different quantities of entity types in each training instance, thereby increasing its adaptability to handle scenarios with varying numbers of entities. The usage of this technique results in an average improvement of over 1.4 points in out-of-domain evaluation, as shown in the Figure 5.

\section{Related Works}
\paragraph{Named Entity Recognition} 
NER is a well-established task in the field of NLP, with numerous applications. Initially, NER models relied on rule-based system \citep{weischedel-etal-1996-progress} that were built using handcrafted algorithms and gazetteers \citep{Mikheev1999NamedER,Nadeau2006UnsupervisedNR,Zamin2011BuildingAC}. However, these models had limitations in terms of scalability and adaptability to new domains or languages. To overcome these issues, machine learning approaches have been proposed \citep{Lafferty2001ConditionalRF}. In the early stages, NER tasks were designed as sequence labeling \citep{huang2015bidirectional, Lample2016NeuralAF,akbik-etal-2018-contextual} where the objective was to predict tagged sequences (e.g., BILOU tags \citep{Ratinov2009DesignCA}). Since then, several paradigm shifts have occurred: span-based approaches treating NER as span classification \citep{Sarawagi2004SemiMarkovCR,fu-etal-2021-spanner,li2021empirical,zaratiana2023filtered}; NER being treated as a question answering problem \citep{Li2019AUM}; and even as a generation task \citep{yan2021unified}.

\paragraph{Zero-shot learning for NER} The advent of large-scale autoregressive models has recently transformed many paradigms in NLP through natural language prompting \citep{Min2022RethinkingTR,Wei2022ChainOT,Qin2023IsCA}. This is also the case for NER \citep{Li2022PromptbasedTE, Ashok2023PromptNERPF,Agrawal2022LargeLM}. Others have fine-tuned these models for tasks to better align their capabilities with the requirements of entity recognition \citep{Cui2021TemplateBasedNE, zhou2023universalner} or information extraction in general \citep{Lou2023UniversalIE, Wang2023InstructUIEMI, Sainz2023GoLLIEAG, Lu2022UnifiedSG}. This is done through-instruction tuning \citep{Mishra2021CrossTaskGV,Wang2022SuperNaturalInstructionsGV,Longpre2023TheFC}.


\begin{figure}[]
    \centering
    \includegraphics[width=1.\columnwidth]{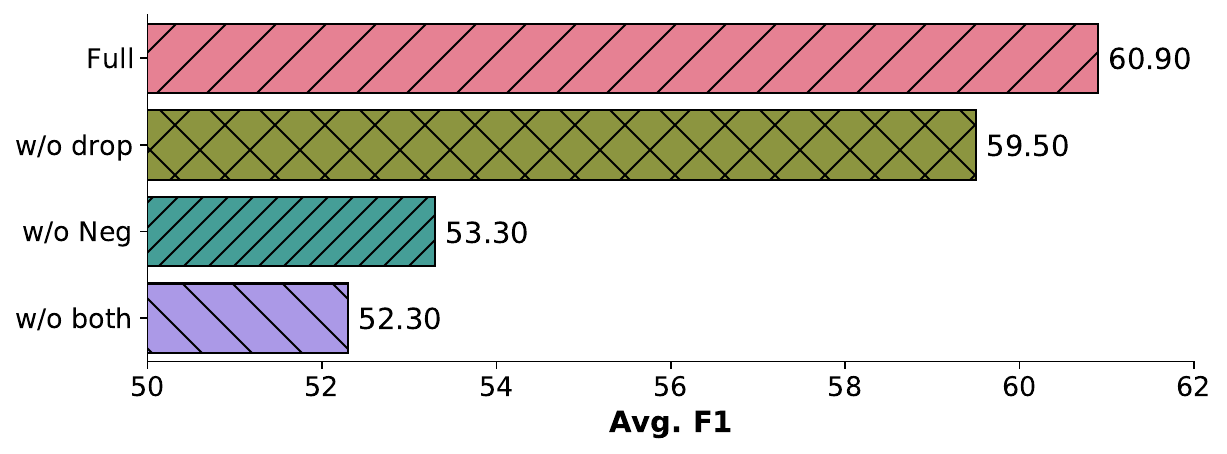}
    \caption{\textbf{Randomly dropping entity types.} We report the results with and without negative entity sampling.}
    \label{fig:throu}
\end{figure}

\section{Conclusion}

In this paper, we introduced GLiNER, a new method for identifying various types of entities in text using bidirectional language models. Our model not only outperforms state-of-the-art Large Language Models like ChatGPT in zero-shot scenarios but also offers a more resource-efficient alternative, crucial for environments with limited computing power. GLiNER is versatile, performing well in multiple languages, including those it wasn't trained on. In future work, we aim to further improve GLiNER's design for enhanced performance and to better adapt it for low-resource languages.

\bibliography{custom}
\bibliographystyle{acl_natbib}

\end{document}